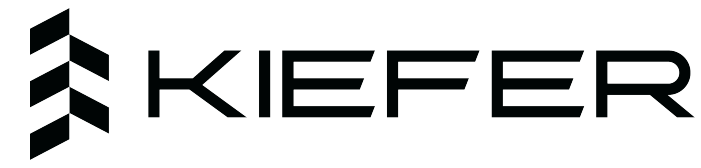

# MORFES: A Benchmark for Productive Inflectional Competence in Modern Greek

**Ioakeim Perros**[1] **Cleopatra Papadopoulou**[1] **Ayoub Kirouane**[1] **Christos Petrocheilos**[1]

[1]Sophea AI, KIEFER SA, Athens, Greece
{i.perros, c.papadopoulou, a.kirouane, c.petrocheilos}@kiefer.gr
models@sophea.ai

Models: huggingface.co/KIEFERSA Dataset: huggingface.co/datasets/KIEFERSA/MORFES

## Abstract

Modern Greek is a richly inflected language, yet the language models built for it are evaluated mainly on factual knowledge, and no benchmark is dedicated to their inflectional competence. We introduce MORFES (Morphological Open-class Recognition-and-Formation Evaluation Suite), a benchmark of 500 expert-verified items that tests the recognition and production of Greek inflected forms, favoring lower-frequency lemmas so that a correct answer reflects the rule rather than a memorized form. We make it publicly available at https://huggingface.co/datasets/KIEFERSA/MORFES. We evaluate a range of open language models on MORFES, situating them within the rapidly scaling open-weight ecosystem from LLaMA to Qwen3, DeepSeek-R1, Magistral, and Kimi K2, where multilingual coverage grows but grammatical competence in morphologically rich languages remains under-measured. Among them, Sophea-Genesis-1, a model we developed and release as open weights at https://huggingface.co/KIEFERSA/Sophea-Genesis-1, leads on inflectional morphology while matching similarly sized models in general capability.

## 1. Introduction

In Modern Greek, every noun, adjective, and verb takes many distinct forms, generated by systematic rules rather than stored one by one, so that a competent speaker produces the correct form of a word even without having encountered that particular form. This distinction matters for a language model because the inflected forms of the language are far too numerous to be acquired by memorization alone: a model that reproduces only the forms well attested in its training data will fail on the many words whose forms occur there rarely or not at all (Goldman et al., 2022), whereas a model that has internalized the rules inflects them correctly. Measuring this competence, as distinct from recall, therefore requires an evaluation directed at words unlikely to have been memorized, so that a correct form is evidence of the rule rather than of prior exposure.

No such evaluation exists for Modern Greek. Its language models are assessed mainly on factual knowledge, and the morphological resources that include Greek either assemble their items automatically rather than by expert curation, as in the SIGMORPHON reinflection shared tasks (Cotterell et al., 2018; McCarthy et al., 2019), or test only subject–verb agreement, as in MultiBLiMP (Jumelet et al., 2026). No benchmark has been dedicated to Greek inflection itself, expert-curated and built to measure productive competence.

We introduce MORFES (also *μορφές*, Greek for 'forms'), the Morphological Open-class Recognition-and-Formation Evaluation Suite, a benchmark of 500 items that measures productive inflectional competence in Modern Greek: the ability to derive the inflected form of a noun, adjective, or verb by rule rather than to retrieve a memorized one. Each item is posed in two modes: recognition, in which the model selects among candidate forms, and production, in which it writes the form itself. The score reflects production and not recognition alone, since identifying a correct form among candidates does not entail the ability to produce it without them. Each item pairs the correct form with near-identical alternatives, so that a correct answer requires the exact form rather than the elimination of an obviously wrong one; coverage traverses the inflectional system of each part of speech; and items favor lower-frequency lemmas, so that a correct form is credited to the rule rather than to recall. Because a Greek specification often admits more than one correct form, any accepted form is credited, though accentuation must be exact; and every item was verified by a native-speaker linguist.

On MORFES, a range of strong open models perform better at recognition than at production scored per item. Among the models evaluated is one we developed, Sophea-Genesis-1, which attains leading performance on inflectional morphology while remaining comparable to the strongest open-weight models of similar size on general-capability benchmarks in English and Greek.

Our contributions are the following:

- **MORFES**, the first benchmark dedicated to Modern Greek inflection: 500 expert-verified items pairing recognition and production, released publicly.
- Its **construction methodology**, in which every design decision derives from the construct of productive competence, including a frequency control that keeps the benchmark a measure of rule rather than recall.
- A **comparative evaluation** of open language models on MORFES, in recognition and production, and on a general-capability panel, including Sophea-Genesis-1, a model we develop and release as open weights.

## 2. Related Work

### 2.1. Morphological Evaluation in NLP

The task MORFES instantiates, producing the correct inflected form from a lemma and a morphosyntactic feature specification, is that of the SIGMORPHON shared tasks on morphological (re)inflection (Cotterell et al., 2018; McCarthy et al., 2019). There, Modern Greek is one language among 103, and its items are produced automatically, by extracting paradigm tables and sampling forms, with no linguistic curation of which forms or phenomena to test. A recent benchmark aimed at language models, IMPACT (Saeed et al., 2025), likewise generates its items synthetically, and the several morphologically rich languages it covers do not include Modern Greek. MORFES, by contrast, is expert-curated: every item is reviewed, corrected, and verified by a linguist; its coverage is directed toward the points of greatest morphological pressure; and item frequency is controlled toward lower-frequency lemmas, so that a correct answer reflects productive competence rather than the recall of a memorized high-frequency form.

Modern Greek is also covered by MultiBLiMP 1.0 (Jumelet et al., 2026), a massively multilingual benchmark of linguistic minimal pairs. Its Greek items, however, target only subject–verb agreement, a morphosyntactic dependency between two words, whereas MORFES tests the productive inflection of the richly inflecting open classes (nouns, adjectives, and verbs) across person, number, gender, case, tense, aspect, mood, and voice. Another line of work evaluates grammar through correction rather than construction: grammatical error correction rewrites a learner's text into a grammatical version of itself, extended to twelve languages, Greek among them, by the MultiGEC-2025 shared task (Masciolini et al., 2025). Its object is the holistic repair of naturally occurring errors in context, scored against corrected reference texts, rather than the elicitation of a specified inflected form.

Recent work by Arnett and Bergen (2025) demonstrates that language models perform worse for morphologically complex languages, finding that the performance gap between agglutinative and fusional languages is most reduced when training datasets are equalized by the "byte-premium", the different encoding efficiencies of different languages and orthographies. A follow-up work (Arnett et al., 2025) expands MorphScore, a tokenizer evaluation metric, to 70 languages, finding that morphological alignment of tokenizers alone does not explain variance in downstream performance. These findings contextualize MORFES: for a fusional language like Greek, productive inflectional competence may depend more on the quality and quantity of language-specific post-training data than on tokenizer design alone.

### 2.2. Evaluation of Modern Greek Language Models

Evaluation of Modern Greek language models has so far centered on factual and subject-matter knowledge rather than on grammar. The native-sourced Greek-MMLU benchmark (Zhang et al., 2026) tests multitask knowledge with multiple-choice questions drawn from Greek academic and professional examinations. The open Greek large language models are assessed on the same footing: Meltemi (Voukoutis et al., 2024) and Krikri (Roussis et al., 2025) report their Greek results on knowledge and general benchmarks, none of which probes grammatical form. What is missing is a benchmark dedicated to Modern Greek inflection itself, curated to test productive competence rather than knowledge, subject–verb agreement, or error correction. MORFES fills that gap.

### 2.3. The Open-Weight Model Landscape

The open-weight ecosystem has scaled rapidly since the release of LLaMA (Touvron et al., 2023), which demonstrated that models trained exclusively on public data can match or exceed proprietary systems at comparable scale. This trajectory has accelerated through Qwen3 (Qwen Team, 2025), which extends multilingual support from 29 to 119 languages and introduces unified thinking/non-thinking modes, and DeepSeek-R1 (DeepSeek-AI, 2025), which shows that reasoning capabilities can be incentivized through pure reinforcement learning without human-labeled reasoning trajectories, with emergent self-reflection and verification patterns that transfer to smaller models via distillation.

Magistral (Mistral AI, 2025) extends this RL paradigm with a method to control the reasoning language of the model, directly relevant to non-English and lower-resource languages, while maintaining multimodal and instruction-following capabilities. Most recently, Kimi K2 (Kimi Team, 2025) pushes the frontier to a 1-trillion-parameter mixture-of-experts architecture with 32B activated parameters, trained on 15.5 trillion tokens with large-scale agentic data synthesis and joint RL, achieving state-of-the-art results among open-source non-thinking models on agentic benchmarks.

Despite this scaling, grammatical competence in morphologically rich, lower-resource languages remains an under-measured axis of evaluation. Benchmark contamination is a growing concern: Ahuja et al. (2024) find that almost all popular multilingual benchmarks show signs of contamination in major LLMs, motivating MORFES's design choices of lemma release for decontamination and preference for lower-frequency lemmas.

# 3. Methodology

## 3.1. Construct

MORFES measures the inflectional-morphological competence of Modern Greek in the open word classes: given a lemma and a grammatical specification, the ability to recognize and to produce the correct inflected form of a verb, noun, or adjective. These three are the richly inflecting open classes, and their inflection is productive: the same rule extends to words the model has never seen. The fourth open class, the adverb, inflects on a single feature (degree) and so falls outside a construct built on rich, productive paradigms. The closed classes, such as articles and pronouns, inflect too, but their paradigms form a small, fixed set that can be memorized, so MORFES excludes them: success on them could not separate a productive rule from a recalled form. A benchmark operationalizes an abstract construct as an observable score, and its construct validity is the degree to which that score reflects the construct and not incidental correlates (Jacobs and Wallach, 2021). The construct MORFES operationalizes is productive inflectional competence: deriving an inflected form by rule rather than retrieving a memorized one. As recommended for benchmarks (Bean et al., 2025), MORFES ties every further design decision to this construct:

- **Recognition and production.** The score reflects production, not recognition alone (Section 3.3).
- **Near-minimal distractors.** Each distractor is a closely related, plausible incorrect form, so that a correct answer reflects command of the exact form required and not the elimination of an obviously wrong option (Section 3.4).
- **Systematic axis coverage.** Traversing each class's inflectional axes rather than sampling for convenience, the score reflects competence across the full inflectional system and not a convenient subset of it (Section 3.4).
- **Lower-frequency lemmas.** A correct form reflects the rule and not the recall of a memorized one (Section 3.5).
- **Every accepted form credited.** Because a Greek cell often has more than one correct form, the score reflects whether the model can produce a correct form and not whether it reproduced the single variant chosen as the reference answer (Section 3.6).

## 3.2. Categories, Formats, and Distribution

MORFES covers the inflection of nouns, adjectives, and verbs, and tests each in two formats.

A **single-form item** asks for one cell of a paradigm, for example the genitive singular of a noun or the third-person plural of a given verb tense. A **full-paradigm item** asks instead for a complete sub-table: for a verb, all six persons of a single combination of voice, tense, aspect, and mood; for a noun, the full set of case and number forms; and for an adjective, that full set for a single given gender, with items distributed across the three genders.

The set is verb-weighted: 300 of the 500 items are verbs, evenly split between the two formats (150 single-form and 150 full-paradigm), with the remaining 200 divided among the noun and adjective formats at 50 items each. A verb has far more distinct inflected forms than a noun or adjective, on the order of a hundred against a couple of dozen for an adjective and only a handful for a noun, so an equal three-way split would under-sample precisely the class that inflects the most.

## 3.3. Recognition and Production

Each item is administered in two modes. In **recognition**, the four candidate forms are shown and the model selects one; in **production**, the same item is posed with no candidates and the model writes the form or forms itself. The two are kept distinct because they probe different abilities: a model can recognize a correct form among alternatives that it could not generate unprompted, so recognition alone would overstate productive competence. Posing both on the same items distinguishes recognizing a correct form from producing it.

### 3.4. Item Selection and Coverage

Every item is built on a single open-class lemma and targets one clearly defined cell of its paradigm. In recognition, each of the three distractors is a near-minimal, plausible incorrect form of the same lemma: most differ from the correct answer by a single grammatical feature (person, number, gender, case, tense, aspect, mood, or voice), while others substitute the inflectional ending of a different class, or introduce an accentuation or spelling error, and some differ by two features. In every case the distractor stays close enough that a correct choice requires the exact form and cannot be reached by eliminating an obviously wrong option. Each lemma appears in only one item, and the requested form is never the lemma's dictionary form (the headword it is listed under), so no item can be answered by repeating the word in the prompt.

Coverage is organized around the inflectional system, not word frequency. Every value of the relevant features is exercised, number, gender, and case for nouns and adjectives, and person, tense, aspect, mood, and voice for verbs, and within each part of speech the items span several inflection classes, not only the most common one. They are weighted toward the points of greatest morphological pressure, the cells whose formation requires a stem change or a stress shift, where the productive rule, and thus a likely error, is most exposed, as in the aorist *παρέλαβα* ('received') of the compound *παραλαμβάνω* ('receive').

The set of inflectional classes to cover was drawn from a Greek reference grammar (Chatzisavvidis and Chatzisavvidou, 2011), used purely as a checklist of what to include (the full inventory, with a plain-language description of each class, appears in Appendix A); the lemmas themselves were chosen independently.

### 3.5. Frequency Control

A benchmark of inflection faces a confound: a model may return the correct form of a common word not by applying the rule but by recalling a form it saw many times in training. To keep the benchmark a test of productive competence rather than recall, MORFES prefers lower-frequency lemmas: the rarer a word, the less plausible that a correct form was memorized rather than derived. This confound is documented. On nonce-word tests, which ask a model to inflect an invented word that has no stored form to recall, large language models show a real-word bias: they return the inflection of a frequent real word in place of the novel form (Weissweiler et al., 2023). Such tests rule out memorization entirely, but only by using words that do not exist in the language. MORFES keeps its items to real, attested Greek and instead limits memorization by preferring lower-frequency lemmas, so that a correct form still reflects the rule and not a familiar stored one. We estimate each lemma's frequency from a large subtitle corpus of Modern Greek, SUBTLEX-GR (Dimitropoulou et al., 2010), and express it on the standard Zipf frequency scale (van Heuven et al., 2014), the base-10 logarithm of a word's frequency per billion tokens, so that higher means more common and a word one point higher is ten times as frequent. We treat a Zipf value of 3 or below as low frequency and preferred, 3 to 4 as transitional, and 4 or above as high frequency and avoided.

The natural estimate is the frequency of a lemma's dictionary form. For nouns and adjectives this is the nominative singular, a common and representative form, so we band them directly on it. For verbs it is the first-person singular present, just one of the verb's many inflected forms and not among its most frequent, so for many common verbs it is itself rare in the corpus and measuring it alone systematically understates the verb's frequency. To measure verb frequency more faithfully, we sum the frequencies of a set of the verb's basic finite forms within a single voice, its present, imperfect, and aorist, rather than relying on the first-person singular present alone, and compute the Zipf value from that sum; restricting the sum to a single voice keeps a verb's active and passive frequencies separate, so each lemma is measured on the forms that actually belong to it. Summing frequency across a word's inflected forms is a long-standing measure in the psycholinguistic study of lexical access (Schreuder and Baayen, 1995); here we apply it within a single voice to fit the Greek verb.

Table 1 gives the resulting distribution. Almost no lemma is high-frequency: none of the 200 nouns and adjectives, and only 8 of the 300 verbs, chiefly high-frequency irregular verbs such as *είμαι* ('be') and *έχω* ('have'): the most irregular verbs of a language are also among its most frequent, so covering their paradigms requires admitting a few high-frequency lemmas. The rest divide between the low and transitional bands. A lemma has no value when the relevant form is unattested

| | Low | Trans. | High | n/a |
|---|---|---|---|---|
| Nouns and adjectives | 86 | 107 | 0 | 7 |
| Verbs | 138 | 147 | 8 | 7 |

Table 1: Frequency-band distribution of the 500 lemmas (Zipf bands: low ≤ 3, transitional 3–4, high ≥ 4). Nouns and adjectives are banded on the frequency of their dictionary form; verbs on the within-voice aggregate. The n/a column counts lemmas with no corpus value.

in the corpus, and a verb when none of its summed forms occur.

### 3.6. Accepted Answers

A grammatical specification can have more than one correct realization in Greek, so each item records a set of accepted forms and credits any of them rather than a single canonical string. The genitive singular of the noun *παύση* (‘cessation’), for example, is credited as both the everyday *παύσης* and the more formal *παύσεως*. The definite article is optional throughout: because the item already fixes case and number, it adds no information, so a form is credited with or without it (both *παύσης* and *της παύσης*). Stress, by contrast, is never optional: it is contrastive in Greek, distinguishing *νόμος* (‘law’) from *νομός* (‘prefecture’), so a form must carry the exact accent to be credited.

### 3.7. Authoring and Integrity

Every item was drafted with the help of a language model[1] and then reviewed, corrected, and verified by a native-speaker linguist; no draft was used unchecked. In particular, while the model proposed candidate items and distractors, the accepted-answer set that defines correctness was authored by the linguist, not the model.

Each item’s lemma is preserved and shared as part of the released benchmark, so that users can hold these lemmas out of a model’s training data and thereby guard against contamination.

## 4. Experiment Results

### 4.1. Setup

We evaluate Sophea-Genesis-1 and a set of open baseline models on MORFES, in both recognition and production. We also report a general-capability panel in English and Greek, to confirm that competence on inflectional morphology does not come at the cost of broad general ability.

Sophea-Genesis-1 is post-trained from Qwen3.6-27B; the training data and recipe are proprietary, and the model is released publicly as open weights. Because we neither assess nor use Sophea-Genesis-1′s vision capabilities, we run its evaluation in text-only mode (vLLM’s `--language-model-only` option). We hold the full benchmark lemma set out of the post-training data that targets inflectional morphology, so Sophea-Genesis-1 receives no morphological supervision on any lemma it is evaluated on.

The baselines fall into three groups: general open models at a comparable scale (Qwen3-32B (Qwen Team, 2025), Gemma-4-31B-it (Gemma Team, 2026), Qwen3.6-27B, and EuroLLM-22B-Instruct (Ramos et al., 2026)); the open Greek models (Llama-Krikri-8B-Instruct (Roussis et al., 2025) and Meltemi-7B-Instruct (Voukoutis et al., 2024)), which are smaller but purpose-built for Greek; and our own general-purpose model Sophea-Titan-1 (27B). For Krikri we report the original release, as the later v1.5 scores lower on the morphology benchmark under our evaluation protocol. The exact Hugging Face repositories for these models are listed in Table 8.

Every model is run under one fixed configuration of the lm-evaluation-harness (Gao et al., 2024) (version 0.4.12) on a vLLM (Kwon et al., 2023) backend, decoding greedily and without reasoning traces (non-thinking mode). Both morphology modes are 0-shot. In production, every model receives the same system instruction:

> *«Είσαι ειδικός στη νεοελληνική γραμματική. Απάντησε μόνο με τους ζητούμενους τύπους, έναν σε κάθε γραμμή, με τη σειρά που ζητούνται, χωρίς γραμματικές ετικέτες και χωρίς προσωπικές αντωνυμίες.»*
>
> English: ‘You are an expert in Modern Greek grammar. Answer only with the requested forms, one per line, in the order asked, without grammatical labels and without subject pronouns.’

followed by the question stem. It prescribes only the output format. Across both morphology and the panel, multiple-choice cells are scored by log-likelihood on the raw completion, without a chat template, while the generative cells apply the model’s chat template.

We report three metrics:

- **Recognition accuracy.** Each of the four candidate forms is scored as a continuation of the question by its log-likelihood, normalized for length so that forms of unequal length compete fairly, and we report the proportion of items on which the highest-scoring form is the correct one.

[1]The drafting model was Claude Opus 4.8, used interactively rather than through a scripted API, so no fixed temperature, token-limit, or system-prompt settings apply; default interactive decoding was used.

| Model | Recognition (%) | Production (%) | |
|---|---|---|---|
| | | per form | per item |
| **Sophea-Genesis-1 (27B)** | **91.2** | **90.0** | **84.0** |
| Qwen3-32B | 77.4 | 52.3 | 35.0 |
| Gemma-4-31B-it | 64.0 | 68.5 | 52.6 |
| Sophea-Titan-1 (27B) | 84.2 | 79.8 | 71.4 |
| Qwen3.6-27B | 81.8 | 72.0 | 63.0 |
| EuroLLM-22B-Instruct | 83.6 | 38.5 | 36.0 |
| Llama-Krikri-8B-Instruct | 81.6 | 30.5 | 30.2 |
| Meltemi-7B-Instruct | 76.8 | 9.4 | 13.6 |

Table 2: MORFES results (500 items, 0-shot): recognition and production accuracy at the form and item level. Best result in each column is in bold.

| Category | Items | Recognition (%) | Production (%) | |
|---|---|---|---|---|
| | | | per form | per item |
| Noun, single form | 50 | 80.0 | 88.0 | 88.0 |
| Noun, full paradigm | 50 | 96.0 | 94.8 | 72.0 |
| Adjective, single form | 50 | 90.0 | 90.0 | 90.0 |
| Adjective, full paradigm | 50 | 98.0 | 95.3 | 84.0 |
| Verb, single form | 150 | 93.3 | 89.3 | 89.3 |
| Verb, full paradigm | 150 | 89.3 | 85.7 | 79.3 |
| **Overall** | **500** | **91.2** | **90.0** | **84.0** |

Table 3: Sophea-Genesis-1 performance on MORFES by category.

- **Production accuracy, per form.** No candidates are shown and the model writes the requested form or forms; we report the proportion of target forms produced correctly, counting a form correct when it exactly matches one of the accepted answers for that cell, accentuation included (some cells admit more than one attested form).
- **Production accuracy, per item.** The same generations scored all-or-nothing: an item counts only if every one of its forms is correct, placing production on the same per-item basis as recognition.

### 4.2. Morphological Competence

Table 2 shows that recognition and production differ in difficulty: all models score well above chance on recognition, whereas their per-item production accuracy is substantially lower. Sophea-Genesis-1 attains the highest accuracy on all three metrics, and its advantage over the baselines is greatest on production, where it scores 84% per item against 71.4% for the strongest baseline.

Table 3 reports Sophea-Genesis-1 by category. Verbs are the most inflectionally complex part of speech, yet production accuracy on verbs is comparable to that on nouns and adjectives, so the model handles them as reliably as the simpler ones.

### 4.3. General Capability

Table 4 reports a general-capability panel spanning knowledge, reading comprehension, and instruction-following in English and Greek. Knowledge is measured by MMLU (Hendrycks et al., 2021) in English, a multiple-choice exam across 57 academic and professional subjects (5-shot, its canonical protocol), and by Greek-MMLU (Zhang et al., 2026) in Greek, a multiple-choice exam natively authored rather than machine-translated from English (0-shot). Reading comprehension is measured by Belebele (Bandarkar et al., 2024) in its English and Greek variants, each item a short passage followed by a four-option question (0-shot). Knowledge and reading are scored by accuracy. Instruction-following is measured by IFEval (Zhou et al., 2023) in English, which checks whether a response satisfies verifiable format constraints such as a length limit or a required keyword (0-shot); it is scored by prompt-level strict accuracy, counting a prompt correct only if every one of its instructions is satisfied.

| Model | MMLU | GreekMMLU | Belebele EN | Belebele EL | IFEval |
|---|---|---|---|---|---|
| **Sophea-Genesis-1 (27B)** | 86.6 | 83.4 | 96.3 | 93.3 | 83.7 |
| Qwen3-32B | 81.9 | 65.1 | 94.6 | 87.2 | 83.4 |
| Gemma-4-31B-it | 63.3 | 81.8 | 96.0 | 92.3 | 90.9 |
| Sophea-Titan-1 (27B) | 86.4 | 83.4 | 95.8 | 92.8 | 77.4 |
| Qwen3.6-27B | 86.6 | 79.5 | 95.8 | 91.8 | 83.5 |
| EuroLLM-22B-Instruct | 68.1 | 72.4 | 82.0 | 78.4 | 71.2 |
| Llama-Krikri-8B-Instruct | 63.7 | 67.3 | 87.7 | 85.0 | 81.1 |
| Meltemi-7B-Instruct | 55.7 | 63.7 | 65.1 | 57.6 | 35.1 |

Table 4: General-capability panel: knowledge (MMLU for English, GreekMMLU for Greek), reading comprehension (Belebele), and instruction-following (IFEval, English only).

Across English and Greek knowledge, reading, and instruction-following, Sophea-Genesis-1 is comparable to the strongest open-weight general-purpose models of similar scale, so its morphological ability does not come at the cost of broad competence.

## 5. Discussion

The results reveal a consistent pattern: while most models achieve reasonable recognition accuracy (selecting the correct form from candidates), their production accuracy drops substantially, particularly at the per-item level where all forms must be correct. This gap between recognition and production is itself informative: it suggests that many models have partial knowledge of Greek morphology sufficient to identify correct forms but insufficient to generate them reliably.

The strong performance of Sophea-Genesis-1 relative to its base model Qwen3.6-27B (84.0% vs. 63.0% per-item production) demonstrates that targeted post-training can substantially improve morphological competence without sacrificing general capability.

The poor production scores of models purpose-built for Greek (Meltemi at 13.6%, Krikri at 30.2%) despite reasonable recognition scores suggest that current Greek-specialized models may rely more on memorized forms than on internalized rules. Such reliance would bound production by what the training data covered, consistent with Arnett and Bergen's (2025) finding that dataset size, scaled by the byte-premium, is the primary driver of performance for morphologically complex languages: these models, though trained on Greek data, may not have sufficient coverage of the long tail of inflected forms.

## 6. Limitations

Every MORFES item was authored and verified by a single native-speaker linguist, so we do not report inter-annotator agreement. However, the correct form is fixed by the inflectional system rather than by annotator judgment, so single authorship is less consequential in this setting than for a task requiring interpretive judgment.

MORFES samples the main inflectional classes rather than exhaustively enumerating every irregular pattern, and its rarest classes carry too few items for a reliable per-class score.

We cannot guarantee that the benchmark's lemmas and their inflected forms are absent from a given model's pretraining data, and public release exposes the benchmark to leakage into future pretraining corpora. Two aspects of the design mitigate this risk: the preference for lower-frequency lemmas (Section 3.5) reduces the plausibility that a correct form reflects memorization, and the release of each item's lemma enables the full lemma set to be excluded from a model's training data, as was done for the morphological supervision of Sophea-Genesis-1 (Section 4.1). The contamination concerns documented by Ahuja et al. (2024) for multilingual benchmarks further motivate this decontamination-by-design approach.

## 7. Conclusion

We presented MORFES, a benchmark for productive inflectional competence in Modern Greek that pairs recog-nition with production and controls for lemma frequency, so that a correct answer reflects the rule rather than recall. Production (writing the requested forms with no candidates shown), scored per item, separates the models more sharply than recognition (selecting among candidate forms). Sophea-Genesis-1, the model we developed, leads on the benchmark while matching open models of similar size in general capability. We release MORFES to make productive inflectional competence in Modern Greek measurable, adding a grammatical benchmark to an evaluation landscape that has centered on factual knowledge.

As the open-weight ecosystem continues to scale (from LLaMA's foundational release through Qwen3′s 119-language coverage, DeepSeek-R1′s RL-incentivized reasoning, Magistral's reasoning-language control, to Kimi K2′s trillion-parameter agentic architecture), MORFES provides a targeted probe for whether this scaling translates into genuine grammatical competence in a morphologically rich language such as Modern Greek, or whether it merely increases the surface coverage of memorized forms.

## Appendix A: Inflectional-Class Inventory

This appendix describes the inflectional classes the benchmark was built to cover. The noun and adjective classes follow the categories of the reference grammar (Chatzisavvidis and Chatzisavvidou, 2011), used as a coverage checklist; the verb classes follow a grouping of our own, defined by the criterion set out in Section A.3. Each table lists, for every class, a plain-language description, an example taken from the benchmark, and the number of items.

### A.1 Nouns

Every noun has a fixed gender (masculine, feminine, or neuter) and changes its ending for grammatical role (case) and for number (singular or plural). A class is identified by the noun's gender and the ending of its dictionary form.

| Description | Example | n |
|---|---|---|
| Feminine nouns ending in -α | γοργόνα | 17 |
| Masculine and feminine nouns ending in -ος | φούρνος | 16 |
| Neuter nouns ending in -ι | σακί | 10 |
| Nouns whose plural adds a syllable, ending in -δες | βοριάς | 9 |
| Feminine nouns ending in -η | βρύση | 9 |
| Masculine nouns ending in -ης, plain plural in -ες | πωλητής | 9 |
| Neuter nouns ending in -ο | σύννεφο | 7 |
| Masculine nouns ending in -ας, plain plural in -ες | γείτονας | 7 |
| Neuter nouns ending in -ος | άλσος | 6 |
| Neuter nouns ending in -μα | δέμα | 5 |
| Masculine nouns ending in -έας | διανομέας | 2 |
| Feminine nouns ending in -ώ | ηχώ | 1 |
| Neuter nouns ending in -υ | δάκρυ | 1 |
| Feminine nouns in -η with formal genitive in -εως | παύση | 1 |

Table 5: The 14 noun classes (100 items), ordered by item count.

### A.2 Adjectives

An adjective takes a different form for each gender, so a class is identified by its three endings: one for the masculine, one for the feminine, and one for the neuter (given here in the dictionary form).

| Description | Example (masc./fem./neut.) | n |
|---|---|---|
| Masc. -ος, fem. -η, neut. -ο; most common | ζεστός / ζεστή / ζεστό | 46 |
| Masc. -ης, fem. -ης, neut. -ες | επιμελής / επιμελής / επιμελές | 21 |
| Masc. -ύς, fem. -ιά/-εία, neut. -ύ | φαρδύς / φαρδιά / φαρδύ | 12 |
| Masc. -ος, fem. -α, neut. -ο | κρύος / κρύα / κρύο | 9 |
| Masc. -ων, fem. -ουσα, neut. -ον; formal | δευτερεύων / -ουσα / -ον | 6 |
| Masc. -άς/-ής, fem. -ού, neut. -άδικο; informal | φωνακλάς / -ού / -άδικο | 3 |
| Masc. -ής, fem. -ιά, neut. -ί; mostly colours | σταχτής / σταχτιά / σταχτί | 1 |
| Masc. -ος, fem. -ια, neut. -ο | φρέσκος / φρέσκια / φρέσκο | 1 |
| Masc. -ης, fem. -α, neut. -ικο; informal | ζαβολιάρης / -α / -ικο | 1 |

Table 6: The 9 adjective classes (100 items), ordered by item count.

### A.3 Verbs

Greek verbs inflect in two voices. In the active voice the subject does the action (*γράφω* "I write"); in the passive voice, marked by the ending *-μαι*, the subject is instead on the receiving end of the action, and can also carry a reflexive ("to oneself") meaning, as in *πλένομαι* "I wash myself / I am washed". Each verb also belongs to one of two conjugations, told apart by which syllable is stressed: first-conjugation verbs are stressed earlier in the word, so the final *-ω* is unstressed (*γράφω*, accented on *γρά*); second-conjugation verbs are stressed on the ending itself (*αγαπώ* "I love", accented on the final *-ώ*). The classes are grouped by how a verb forms its simple past, the aorist, because the past stem is the least predictable part of the Greek verb and the main source of its irregularity.

| **Description** | **Example (pres. → aor.)** | **n** |
|---|---|---|
| Passive verbs, aorist in -θηκα | λύνομαι → λύθηκα | 38 |
| 2nd-conj. active, aorist in -ησα | ευδοκιμώ → ευδοκίμησα | 34 |
| Active in -ίζω, aorist in -ισα | θερίζω → θέρισα | 25 |
| Active in -νω, aorist in -σα | ζυμώνω → ζύμωσα | 25 |
| Passive verbs, aorist in -στηκα | χτενίζομαι → χτενίστηκα | 21 |
| Active in -εύω, aorist in -εψα | κουρεύω → κούρεψα | 18 |
| Active, aorist in -ξα (no augment) | αρμέγω → άρμεξα | 18 |
| "Become X" in -αίνω/-ύνω, aor. -υνα | λεπταίνω → λέπτυνα | 13 |
| Compound, internal augment, aor. -ψα | προτρέπω → προέτρεψα | 13 |
| Compound, internal augment, aor. -σα | εκφράζω → εξέφρασα | 11 |
| Compounds of βάλλω, aor. -έβαλα | διαβάλλω → διέβαλα | 10 |
| Passive verbs, aorist in -τηκα | χαϊδεύομαι → χαϊδεύτηκα | 9 |
| Active in -άζω, aorist in -ασα | βράζω → έβρασα | 8 |
| Active, aorist in -ψα (no augment) | τρίβω → έτριψα | 7 |
| Learned verbs, formal aorist in -ην | μεταβαίνω → μετέβην | 6 |
| Compounds of άγω, aor. -ήγαγα | εξάγω → εξήγαγα | 5 |
| Compounds of (σ)τέλλω, aor. -έ(σ)τειλα | αναστέλλω → ανέστειλα | 5 |
| Irregular (different root in aorist) | λέω → είπα | 5 |
| Active in -ρνω/-λνω, stem vowel change | σπέρνω → έσπειρα | 4 |
| Compounds of λαμβάνω, aor. -έλαβα | παραλαμβάνω → παρέλαβα | 4 |
| Passive verbs, aorist in -χτηκα | τυλίγομαι → τυλίχτηκα | 4 |
| Active in -θω, aorist in -σα | αλέθω → άλεσα | 3 |
| Active, aorist in -υσα | μηνύω → μήνυσα | 3 |
| Active, aorist in -εσα | επαινώ → επαίνεσα | 2 |
| Active, aorist in -ηξα | βογκώ → βόγκηξα | 2 |
| Compound, internal augment, aor. -ξα | εκλέγω → εξέλεξα | 2 |
| Learned in -έω, aorist in -ευσα | πλέω → έπλευσα | 2 |
| A verb with no aorist at all | φλέγομαι | 1 |
| Active, aorist in -αξα | βαστώ → βάσταξα | 1 |
| Compounds of νέμω, aor. -ειμα | διανέμω → διένειμα | 1 |

Table 7: The 30 verb classes (300 items), ordered by item count.

## Appendix B: Model Repositories

The exact Hugging Face repository for each model compared against Sophea-Genesis-1 is provided in Table 8.

| **Model** | **Hugging Face Repository** |
|---|---|
| Sophea-Genesis-1 | KIEFERSA/Sophea-Genesis-1 |
| Sophea-Titan-1 | KIEFERSA/Sophea-Titan-1 |
| Qwen3-32B | Qwen/Qwen3-32B |
| Gemma-4-31B-it | google/gemma-4-31B-it |
| Qwen3.6-27B | Qwen/Qwen3.6-27B |
| EuroLLM-22B-Instruct | utter-project/EuroLLM-22B-Instruct-2512 |
| Llama-Krikri-8B-Instruct | ilsp/Llama-Krikri-8B-Instruct |
| Meltemi-7B-Instruct | ilsp/Meltemi-7B-Instruct-v1.5 |

Table 8: Hugging Face repositories for all evaluated models.